\documentclass[runningheads]{llncs}
\usepackage{graphicx,amsmath}
\usepackage{amssymb}
\usepackage{latexsym}
\begin{document}
\title{Multi-level Chaotic Maps for 3D Textured Model Encryption}
%
%\titlerunning{Abbreviated paper title}
% If the paper title is too long for the running head, you can set
% an abbreviated paper title here
%
\author{Xin Jin\inst{1,2}\orcidID{0000-0003-3873-1653} \and
Shuyun Zhu\inst{1} \and
Le Wu\inst{1} \and
\\Geng Zhao\inst{1}  \and
Xiaodong Li\inst{1,*} \and
Quan Zhou\inst{3,4}
Huimin Lu\inst{5}}
\authorrunning{X. Jin et al.}
% First names are abbreviated in the running head.
% If there are more than two authors, 'et al.' is used.
%
\institute{Department of Cyber Security, Beijing Electronic Science and Technology Institute, Beijing, China \and
CETC Big Data Research Institute Co.,Ltd., Guiyang, Guizhou, China \and
Nanjing University of Posts \& Telecommunications, Nanjing, China. \and
Nanjing University, Nanjing, China \and
Department of Mechanical and Control Engineering, Kyushu Institute of Technology, Kitakyushu, Japan \\
\email{*Corresponding author: lxd@besti.edu.cn}}
\maketitle              % typeset the header of the contribution
\begin{abstract}
With rapid progress of Virtual Reality and Augmented Reality technologies, 3D contents are the next widespread media in many applications. Thus, the protection of 3D models is primarily important. Encryption of 3D models is essential to maintain confidentiality. Previous work on encryption of 3D surface model often consider the point clouds, the meshes and the textures individually. In this work, a multi-level chaotic maps models for 3D textured encryption was presented by observing the different contributions for recognizing cipher 3D models between vertices (point cloud), polygons and textures. For vertices which make main contribution for recognizing, we use high level 3D Lu chaotic map to encrypt them. For polygons and textures which make relatively smaller contributions for recognizing, we use 2D Arnold's cat map and 1D Logistic map to encrypt them, respectively. The experimental results show that our method can get similar performance with the other method use the same high level chaotic map for point cloud, polygons and textures, while we use less time. Besides, our method can resist more method of attacks such as statistic attack, brute-force attack, correlation attack. 

\end{abstract}
\section{Introduction}
\label{sec:intro}

Nowadays, more and more images and videos are flooded in our daily lives. In addition to images and video, 3D models are beginning using 3D modeling and 3D printing. Certain apps on smartphones, such as Autodesk 123D Catch, which let users to take a themed photo from various views and upload all the photos to the Autodesk cloud server. The 123D service on the cloud server will then return the 3D model of the theme to the user. Desktop software like Google Sketchup can also easily edit 3D models. 3D models gradually enter our daily life.

Virtual reality technology has become an increasingly popular topic in the industry, and a lot of 3D models are needed to build virtual worlds. The virtual reality and augmented reality market is expected to reach \$1.06 billion in 2018, with a compound annual growth rate (CAGR) of 15.18 \% from 2013 to 2018. The government is scanning the entire city's 3-d virtual city model with laser scanners and multi-view cameras. In order to achieve a high level of security, integrity, confidentiality and protection against unauthorized access to sensitive information, 3D content encryption technology is needed. 3D content is stored or transmitted through unsafe channels.

3D digitized objects are defined by two types of 3D content: 3D solid model and 3D surface (shell/boundary) model. The solid model defines the volume of the physical object represented, while the surface model represents the surface, not the volume. Rey solves the encryption problem of 3D entity model in \cite {ReyHAIS2015}.

\begin{figure*}
\centering
\includegraphics[width=\textwidth]{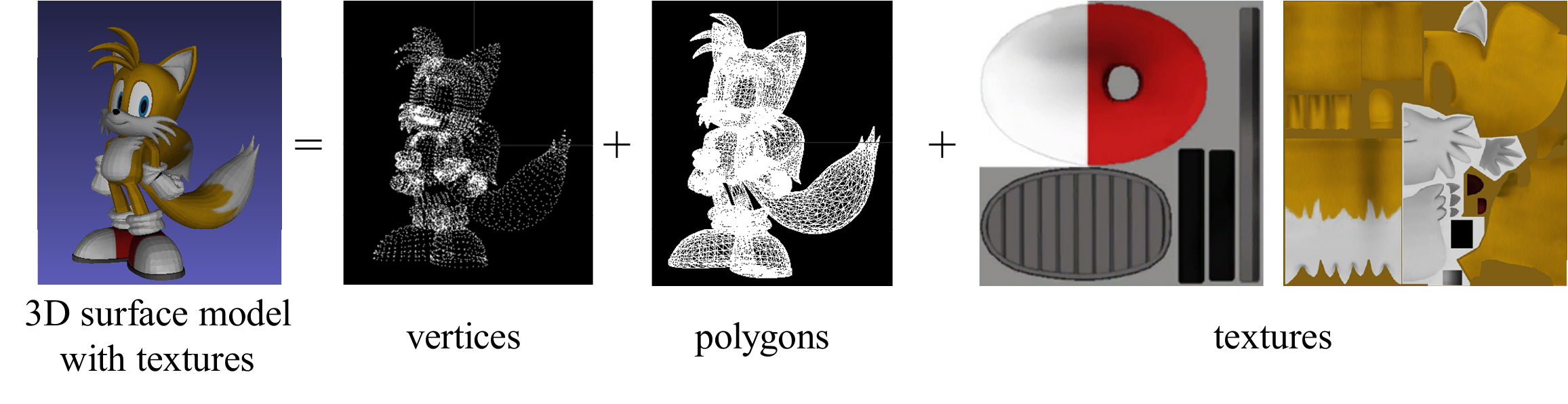}
\caption{The basic composition of a 3D surface model consists of vertices, polygons, and textures.}
\label{fig:3dSurfaceModel}
\end{figure*}

This paper focuses on the encryption of three-dimensional surface models with texture. The direction of the current method is usually considered point cloud encryption \cite{JolfaeiTIFS2015,JinPCM2016}, grid \cite{EluardCORESA2013} and textures \cite{JolfaeiPsivt2015}. Complete 3D surface models usually contain vertices, polygons and textures, as shown in \ref{fig:3dSurfaceModel}.

We have the observation that the vertices, the polygons and the textures make different levels of contributions to recognize a 3D surface model. %as shown in Figure \ref{fig:VPT}. Three main observations are list below.

%\begin{itemize}
%\item[(1)] As long as the encrypted vertices are not attacked, whether or not the encrypted polygons or textures are attacked, one almost cannot recognition the attacked results. See (a)(b)(c)(d) in each example. 
%
%\item[(2)] The polygons and the textures have no such properties of vertices in (1). In (a)(c)(e)(f), the encrypted polygons are not attacked. In (a)(b)(e)(g), the encrypted textures are not attacked.
%
%\item[(3)] Once the encrypted vertices are attacked, as long as the encrypted polygons are not attacked, one can only recognize the basic shapes of the plain 3D models in the attacked results, no matter whether or not the encrypted textures are attacked (e)(f). 
%
%\item[(4)] Once the encrypted vertices and the polygons are attacked, one can easily recognize the plain 3D models in the attacked results, no matter whether or not the encrypted textures are attacked (g)(h). 
%\end{itemize}
%
%
%\begin{figure*}
%\centering
%\includegraphics[width=\textwidth]{VPT}
%\caption{Various attacks on the encrypted 3D surface models with textures. \emph{V} for vertices, \emph{P} for polygons, \emph{t} for textures. The green \checkmark means successful attack, i.e. correct decryption. The red $\times$ means unsuccessful attack, i.e. incorrect decryption. Two examples (\emph{basket} and \emph{radio}) are shown to illustrate the different levels of contributions of vertices, polygons and textures to recognize a 3D surface model with textures.}
%\label{fig:VPT}
%\end{figure*}

Therefore, in this work, we propose a 3D texture model encryption method based on multi-level chaotic mapping of vertices (point clouds), polygons and textures. For vertices mainly used for identification, we use advanced 3D Lu chaotic mapping to encrypt them. For polygons and textures with relatively small contributions to recognition, we used cat map of 2D Arnold and 1D Logistic map to encrypt them, respectively. Experimental results show that our method achieves similar performance in the same high-level chaotic mapping of vertices, polygons and textures with less time consuming. In addition, our method can resist many kinds of attacks, such as brute force attack, statistical attack and related attack.

%\textbf{Organization}. We organize the rest of our paper as following: In section \ref{sec:previous}, we make a brief review of related work in this field. We give the preliminaries in Section \ref{sec:Preliminaries}. The core method is presented in Section \ref{sec:method}. The simulation results are shown in Section \ref{sec:simulation}. We make security and performance analysis in Section \ref{sec:analysis}. At last we give conclusions and discuss future work in Section \ref{sec:conclusion}.

\section{Previous work}
\label{sec:previous}

The special properties of chaos\cite{VermaJIPS2013}, such as sensitivity, pseudorandomness and ergodicity to initial conditions and system parameters, make chaos dynamics a promising alternative to traditional encryption algorithms. Intrinsic attributes directly relate them to obfuscated and diffuse cryptographic characteristics, which are referenced in Shannon's work by \cite{Shannon1949}.

In this respect, the current approach only considers solid models \cite{ReyHAIS2015}, point cloud models \cite{JolfaeiTIFS2015,JinPCM2016}, meshes \cite{EluardCORESA2013}, and textures \cite{JolfaeiPsivt2015}. This paper examines the texture encryption of the most complex 3D surface models.

To the best of our knowledge, the most similar work to ours is Jin el al. \cite{JinArXiv2017}, in which they also propose a method for the encryption of 3D surface models with textures. However, they use the same high level chaotic maps for vertices, polygons and textures. This makes the time consumption of their method is high. Based on our observations in Section \ref{sec:intro}, we choose to use a multi-level (hierarchical) way of using high level chaotic for vertices, middle one for polygons and low one for textures.

\section{Preliminaries}
\label{sec:Preliminaries}
The three levels of chaotic maps we leveraged in this work are 1D logistic map, 2D Arnold's cat map and 3D Lu map. HBecause of its high complexity, high-dimensional chaotic systems are more reliable in designing secure image encryption schemes. The cryptosystem based on low-dimensional chaotic mapping has shortcomings such as short cycle and small key space \cite{Zhen2015,JinCAC2015,JinATIS2015,JinWCSP2016,JinMTA2017}. However, the low-dimensional chaotic maps require less computational cost than that of the high-dimensional chaotic maps. Thus in this paper, we combine the high security of high-dimensional chaotic maps and the high speed of low-dimensional chaotic maps.

\subsection{1D Logistic Map} 

The basic formula of 1D chaotic encryption is as follows:

\begin{eqnarray}
\begin{split}
& x_{n+1} = \mu x_n(1-x_n)\\
& 3.569945672... < \mu \leq 4, 0 \leq x_n \leq 1\\
& n =0,1,2,... 
\end{split}
\label{eq:logistic}
\end{eqnarray}
When $ 3.569945672... < \mu \leq 4, 0 \leq x_0 \leq 1$, the system is in chaotic state.

\subsection{2D Arnold's Cat Map} 

Cat mapping is a chaotic system that iterates or evolves the plaintext as the initial value of the chaotic system to achieve the effect of scrambling the plaintext.

\begin{eqnarray}
\begin{bmatrix}
 &X' \\ 
 &Y' 
\end{bmatrix}
=
\begin{bmatrix}
 &1 &p\\ 
 &q &p*q+1
\end{bmatrix}
*
\begin{bmatrix}
 &X \\ 
 &Y 
\end{bmatrix}
\operatorname{mod}U
\label{eq:Arnold1}
\end{eqnarray}

\begin{eqnarray}
\begin{bmatrix}
 &X \\ 
 &Y 
\end{bmatrix}
=
\begin{bmatrix}
 &1 &p\\ 
 &q &p*q+1
\end{bmatrix}
^{-1}
*
\begin{bmatrix}
 &X' \\ 
 &Y' 
\end{bmatrix}
\operatorname{mod}U,
\label{eq:Arnold2}
\end{eqnarray}
where $p$ and $q$ represent the positive secret keys. $(X,Y)$ is the original 2D variables. $(X',Y')$ is the new values of $(X,Y)$. $U$ is the upper bounds of values of $X$ and $Y$.

\subsection{3D Lu Map}
The Lu map is a 3D chaotic map. It is described by  Eq. \ref{eq:Lu}

\begin{eqnarray}
%\begin{split}
\begin{cases}
 & \dot{x} = a(y-x)\\ 
 & \dot{y} = -xz+cy,\\
 & \dot{z} = xy-bz 
\end{cases}
%\end{split}
\label{eq:Lu}
\end{eqnarray}
where $(x,y,z)$ are the system trace. $(a,b,c)$ are the system parameters. When $a=36, b=3, c=20$, the system contain a strange attractor and is in chaotic state.

\section{Multi-level 3D Model Encryption}
\label{sec:method}

In this section, we describe the proposed encryption method for a textured 3D surface model. First, the 3D texture model is decomposed into vertices, polygons, and textures. Then, the three parts are encrypted by using three-dimensional Lu map, two-dimensional Arnold graph and one-dimensional Logistic graph, respectively, as described in Section \ref{sec:Preliminaries}. Finally, the encrypted vertices, polygons, and textures are combined into an encrypted 3D texture model.

\begin{figure*}
\centering
\includegraphics[width=0.9\textwidth]{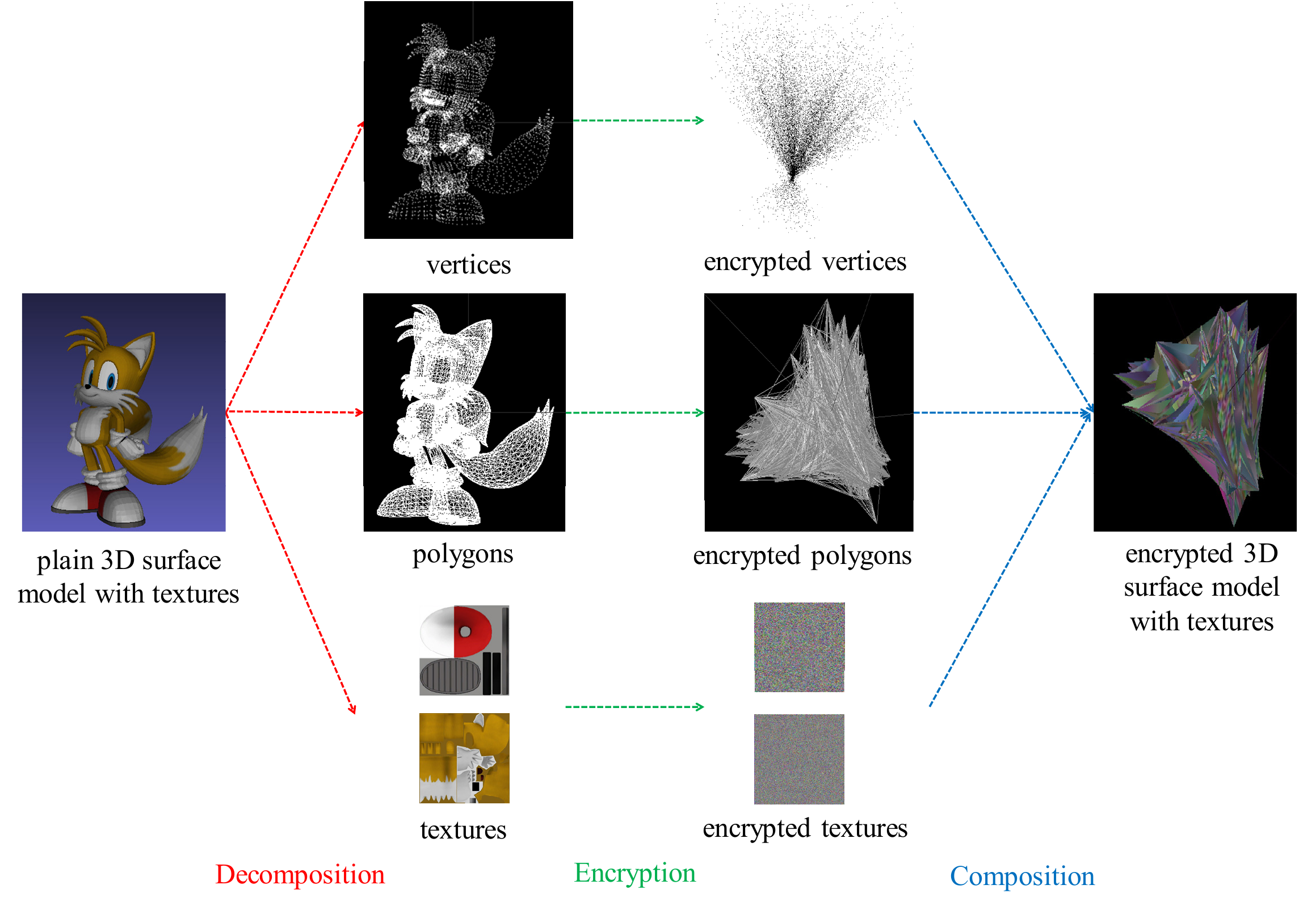}
\caption{Our presented 3-dim textured encryption model. The decryption method is the inverse version of the encryption method.}
\label{fig:overview}
\end{figure*}

\subsection{Vertices Encryption}
The vertices in the 3D texture model are expressed by the following triple list:

\begin{eqnarray}
V = \{(X_1,Y_1,Z_1),...,(X_N,Y_N,Z_N)\},
\end{eqnarray}
where $(X_i,Y_i,Z_i)$ is the 3D coordinate of a vertex. $N$ is the number of the vertices.  We use the 3D Lu map defined in Eq. \ref{eq:Lu} to produce a random vector with dimensions of $3N$:

\begin{small}
\begin{equation}
LV = \{(LV_1,LV_2,LV_3),...,(LV_{3N-2},LV_{3N-1},LV_{3N})\}.
\end{equation}
\end{small}
Then we make element by element product of $V$ and $LV$:

%\begin{tiny}
\begin{footnotesize}
\begin{equation}
\begin{split}
& VLV = \\
& \{(X_1LV_1,Y_1LV_2,Z_1LV_3),...,(X_NLV_{3N-2},Y_NLV_{3N-1},Z_NLV_{3N})\}.
\end{split}
\end{equation}
\end{footnotesize}
%\end{tiny}
The new vector VLV contains the new coordinates of the original 3D vertex:

\begin{small}
\begin{equation}
\begin{split}
&(X_i,Y_i,Z_i) \rightarrow \\ &(X_iLV_{3(i-1)},Y_iLV_{3(i-1)+1},Z_iLV_{3(i-1)+2}), 1\leq i\leq N
\end{split}
\end{equation}
\end{small}

\subsection{Polygons Encryption}
The polygons (taking the triangle as an example) in a 3D textured model are in the form of a list of triplets:

\begin{eqnarray}
P = \{(A_1,B_1,C_1),...,(A_i,B_i,C_i),...,(A_M,B_M,C_M)\},
\end{eqnarray}
where $(A_i,B_i,C_i)$ represents the 3 vertices of a triangle in the form of the indices of vertices. $1\leq i\leq M, 1\leq A_i, B_i,C_i\leq N$. $N$ is the number of the vertices. We use the 2D Arnold's Cat map defined in Eq. \ref{eq:Arnold1} and \ref{eq:Arnold2} to produce a random vector with dimensions of $3M$:

\begin{eqnarray}%\footnotesize
LP = \{(LP_1,LP_2,LP_3),...,(LP_{3M-2},LP_{3M-1},LP_{3M})\}.
\end{eqnarray}
We make the element to element correspondences between $P$ and $LP$:

\begin{eqnarray}
\label{eq:correspond}
\begin{cases}
A_i \longleftrightarrow LP_{3(i-1)+1} \\
B_i \longleftrightarrow LP_{3(i-1)+2} \\
C_i \longleftrightarrow LP_{3(i-1)+3}
\end{cases}
\end{eqnarray}
Then we make ascending sort of $LP$. The sorted $LP$ is denoted as $LP^{sort}$. According to the new order in $LP^{sort}$, we reorder the element in $P$ using the correspondences described in Eq. \ref{eq:correspond}. The vector with new order of $P$ is denoted as $P'$.
\begin{eqnarray}
P' = \{(A'_1,B'_1,C'_1),...,(A'_i,B'_i,C'_i),...,(A'_M,B'_M,C'_M)\},
\end{eqnarray}
where $(A'_i,B'_i,C'_i)$ is the new triangle of the encrypted 3D model.

\subsection{Textures Encryption}
In texture 3D model, texture is represented as 2D image with corresponding texture coordinates. We use 1D logic mapping based image encryption method and DNA coding \cite{JinCAC2015} to encrypt the texture image. We first divide the texture image into RGB channels. Each channel of the texture image is then encoded by DNA coding. We then used the 1D logical mapping to generate a random matrix of texture images of the same size, and added it to the coding results using DNA additions. After that, another random matrix with the same size texture image is generated by 1D logical mapping and converted into a binary matrix with a threshold of $0.5 $. Then, when the corresponding value in the second random matrix is 1, the DNA addition result is converted to the DNA complement result. The final step is to decode the DNA to get an 8-bit encryption result.

\section{Simulation Performance}
\label{sec:simulation}
\begin{figure*}
\centering
\includegraphics[width=0.9\textwidth]{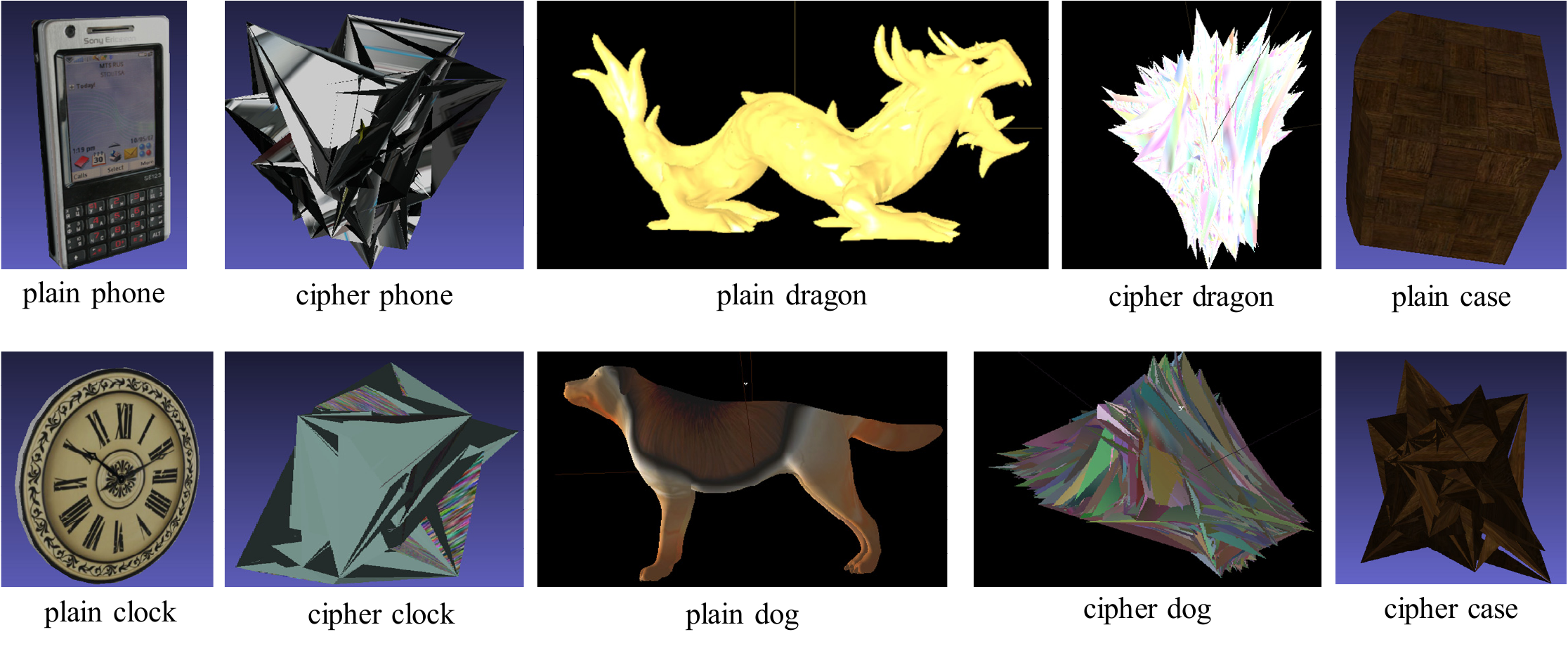}
\caption{The simulation results. we test our method on 3D models with various contents.}
\label{fig:results}
\end{figure*}

We use plenty of 3D textured models to test our method, as shown in Fig. \ref{fig:results}, with the secret keys shown in Table \ref{tb:keys}.

\begin{figure*}
\centering
\includegraphics[width=0.85\textwidth]{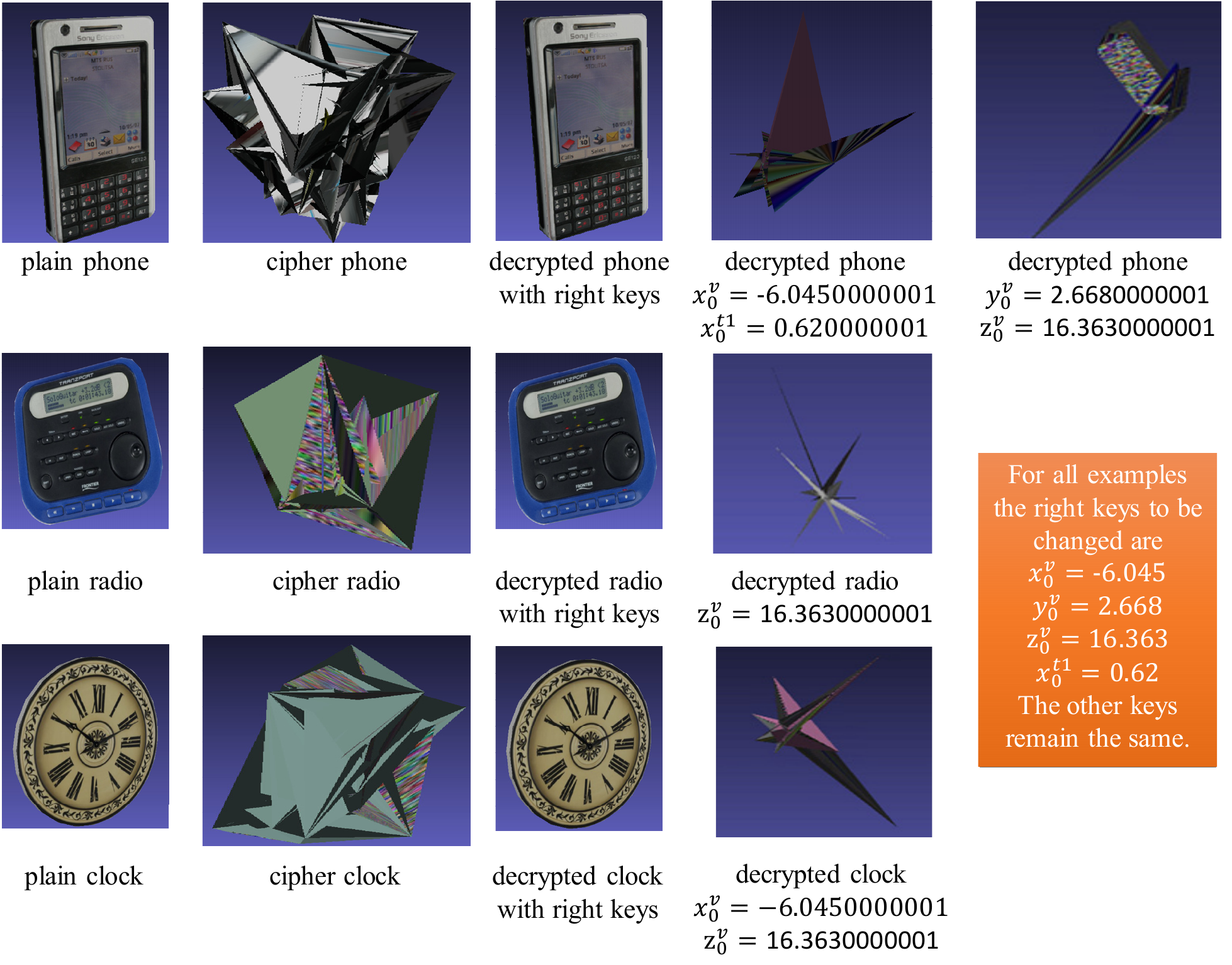}
\caption{Decrypted with wrong key. We slightly change the key and get the wrong decrypted result.}
\label{fig:key}
\end{figure*}

\begin{figure*}
\centering
\includegraphics[width=0.85\textwidth]{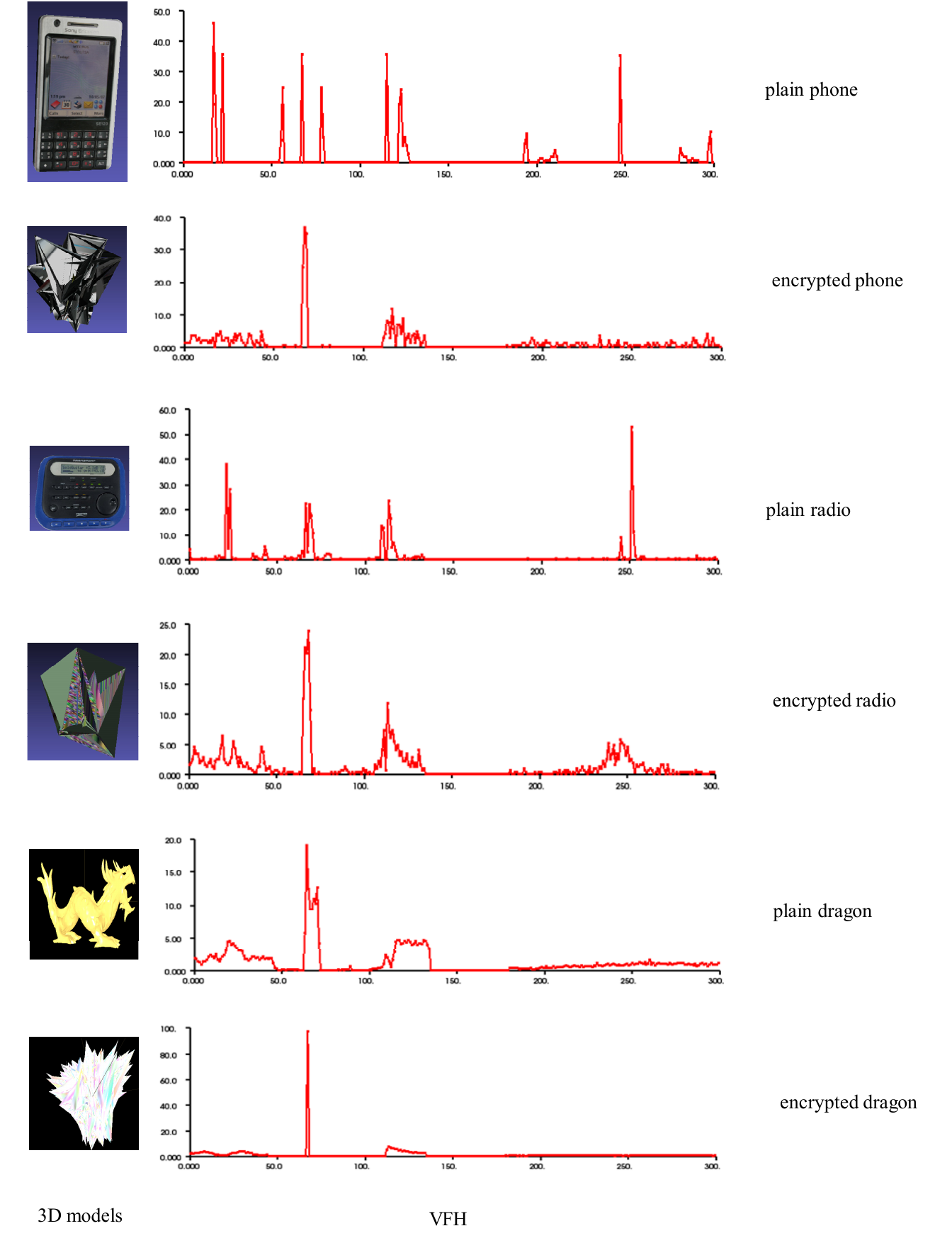}
\caption{The Viewpoint Feature Histogram (VFH) of 3D textured models before and after encryption.}
\label{fig:VFH}
\end{figure*}

\begin{figure*}
\centering
\includegraphics[width=0.9\textwidth]{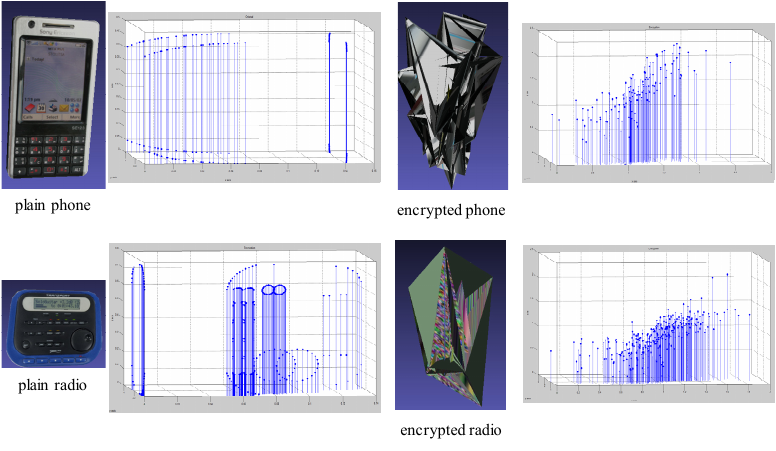}
\caption{The distribution of occupied positions per z-column of the 3D textured models before and after encryption.}
\label{fig:DOP}
\end{figure*}

\begin{table*}[!t]%\footnotesize
\renewcommand{\arraystretch}{1}
\caption{The secret keys of the 3D Lu maps in Eq.\ref{eq:Lu}. In the vertices encryption phases, $a=36, b=3, c=20$.}
\label{tb:keys}
\centering
\begin{tabular}{|c|c|}
\hline
Encryption Phases&	Keys\\
\hline
\hline
vertices encryption&	$x_0^{v}=-6.045, y_0^{v}=2.668, z_0^{v}=16.363$\\
\hline
polygons encryption&	$p=1,q=1$\\
\hline
texture encryption&		$x_0^{t1}=0.62, \mu^{t1}=3.99, x_0^{t2}=0.26, \mu^{t2}=3.9$\\
\hline
\end{tabular}
\end{table*}

In our approach, we use 1 Lu map, 1 Arnold's cat map, and 2 Logistic maps. Texture encryption contains 2 Logistic maps. Three-dimensional texture models with different content were tested. All encryption results can be correctly decrypted to the original pure 3D model with the correct secret key. The simulation results show that the simulation results are satisfactory.

\section{Security and Performance Analysis}
\label{sec:analysis}
Well-designed 3D model encryption schemes should be able to resist various attacks, such as violent attacks and statistical attacks. In this section, we will analyze the security of the proposed encryption method.

\subsection{Resistance to the brute-force Attack}
\subsubsection{Key Space}

The key space of the 3D model encryption scheme should be large enough to resist violent attacks, otherwise it will be broken down by exhaustive search to obtain the secret key in a limited amount of time. In our encryption method, we have the keyspace of 9 key values shown in table \ref{tb:keyspace}.

\begin{table*}[!t]%\footnotesize
\renewcommand{\arraystretch}{1}
\caption{The key spaces.}
\label{tb:keyspace}
\centering
\begin{tabular}{|c|c|}
\hline
Chaotic Maps&	Key Spaces\\
\hline
\hline
3D Lu&	$-40<x_0^{v}<50, -100<y_0^{v}<80, 0<z_0^{v}<140$\\
\hline
2D Arnold' cat map&	$p,q~are~positive~integers$\\
\hline
1D logistic map&	$3.569945672...< \mu^{t1}, \mu^{t2} \leq 4, x_0^{t1}, x_0^{t2} \in [0,1]$\\
\hline
\end{tabular}
\end{table*}

The precision of 64-bit double data is $10^{-15}$, thus the key space is about $(10^{15})^{9}=10^{135} \approx 2^{449}$, which is much lager than the max key space ($2^{256}$) of practical symmetric encryption of the AES \cite{BogdanovAsiacrypt2011}. In our practice, we use double data to simulate large integers for $p$ and $q$ in Table \ref{tb:keys}.   Our key space is large enough to resist brute-force attack.

\subsubsection{Sensitivity of Secret Key}
The chaotic systems are extremely sensitive to the system parameter and initial value. A light difference can lead to the decryption failure. To test the secret key sensitivity of the 3D model encryption scheme.
% we change the secret key as shown in Table \ref{tb:keysen}.

%\begin{table}[!t]\footnotesize
%\renewcommand{\arraystretch}{1}
%\caption{Slightly change the key values.}
%\label{tb:keysen}
%\centering
%\begin{tabular}{|c|c|c|}
%\hline
%Secret Keys&	Original Values&	Novel Values\\
%\hline
%\hline
%$x_0^{v}$&	-6.045&	-6.0450000001\\
%\hline
%$x_0^{t1}$&	0.62&	0.620000001\\
%\hline
%\end{tabular}
%\end{table}

We use the modified key to decrypt the encrypted 3D surface model while the other keys remain unchanged. The decryption result is shown in Fig. \ref{fig:key}. We can see that the decrypted 3D model is completely different from the original plain 3D model. The test results for the other key are similar. Experiments show that the 3D model encryption scheme is very sensitive to keys and has strong resistance to exhaustive attacks.

\subsection{Resistance to the Statistic Attack}
\subsubsection{The Histogram Analysis}
For the vertex, the viewpoint feature histogram (VFH) is the representation of point clusters for cluster identification and 6 DOF pose estimation problems. We use VFH to evaluate our 3D vertex encryption. As shown in figure \ref{fig:VFH}, the VFH of our method's encryption results is completely different from the VFH of the original 3D model, which makes statistical attacks impossible.

\subsubsection{Distribution of Occupied Positions}
We further analyse the occupied positions of the 3D vertices. As defined in \cite{ReyHAIS2015}, we compute the occupied position per $x$-column, $y$-column and $z$-column of a 3D lattice $Z=(z_{ijk})$.

The matrix is very different for normal 3D vertices and encrypted 3D vertices.
The occupied position of each z column in the plane 3D vertex and the corresponding encrypted 3D vertex are shown in Figure \ref{fig:DOP}. The positions of the occupation were distributed widely. In the case of ordinary 3D vertices, some clusters appear, while in the case of encrypted 3D vertices, the distribution appears to be uniform.

\subsection{The Speed of the Encryption and Decryption}

Our 3D surface model encryption scheme is implemented on personal computers by Matlab with AMD A10 PRO-7800B, 12 computer cores 4C + 8G 3.4ghz and 4.00g RAM. Time spent encrypting and decrypting 3D models with different number of vertices. The larger the size of the 3D model, the more time it takes to encrypt and decrypt it. When we implemented the migration to other tool environments (such as C/C ++) in Matlab 2015a, the speed was faster and could meet the actual needs.

\section{Conclusions}
\label{sec:conclusion}
In this paper, we propose a hierarchical multi-level encryption scheme based on multiple dimensional chaotic maps for 3D space models with textures. Our method is based on the observations that the vertices make more contribution to recognizing a 3D model than polygons and textures. The hierarchical scheme reduces the time consumption of encryption and decryption compared with the methods of Jin et al. \cite{JinArXiv2017}, who make equal level encryption on vertices, polygons and textures. While our encryption results and the key sensitivity are nearly the same to those of \cite{JinArXiv2017}. In the future work, we will try to find more hierarchical encryption schemes for 3D models and rely on more sophisticated psychological experiments on 3D model recognition.

%\appendices
%\section{Proof of the First Zonklar Equation}
%Appendix one text goes here.

% you can choose not to have a title for an appendix
% if you want by leaving the argument blank
%\section{}
%Appendix two text goes here.

% use section* for acknowledgment
\section*{Acknowledgement}

This work is partially supported by the National Natural Science Foundation of China (Grant Nos. 61772047, 61772513), the Science and Technology Project of the State Archives Administrator (Grant No. 2015-B-10), and the Fundamental Research Funds for the Central Universities (Grant No. 328201803, 328201801).

%
% ---- Bibliography ----
%
% BibTeX users should specify bibliography style 'splncs04'.
% References will then be sorted and formatted in the correct style.
%
% \bibliographystyle{splncs04}
% \bibliography{mybibliography}

\begin{thebibliography}{10}
\providecommand{\url}[1]{\texttt{#1}}
\providecommand{\urlprefix}{URL }
\providecommand{\doi}[1]{https://doi.org/#1}

\bibitem{BogdanovAsiacrypt2011}
Bogdanov, A., Khovratovich, D., Rechberger, C.: Biclique cryptanalysis of the
  full {AES}. In: Advances in Cryptology - {ASIACRYPT} 2011 - 17th
  International Conference on the Theory and Application of Cryptology and
  Information Security, Seoul, South Korea, December 4-8, 2011. Proceedings.
  pp. 344--371 (2011). \doi{10.1007/978-3-642-25385-0_19},
  \url{https://doi.org/10.1007/978-3-642-25385-0_19}

\bibitem{JinATIS2015}
Jin, X., Chen, Y., Ge, S., Zhang, K., Li, X., Li, Y., Liu, Y., Guo, K., Tian,
  Y., Zhao, G., Zhang, X., Wang, Z.: Color image encryption in cie l*a*b*
  space. In: International Conference on Applications and Techniques in
  Information Security, Beijing, China, November 4-6, 2015. pp. 74--85 (2015)

\bibitem{JinCAC2015}
Jin, X., Tian, Y., Song, C., Wei, G., Li, X., Zhao, G., Wang, H.: An invertible
  and anti-chosen plaintext attack image encryption method based on dna
  encoding and chaotic mapping. In: 2015 Chinese Automation Congress (CAC). pp.
  1159--1164 (Nov 2015). \doi{10.1109/CAC.2015.7382673}

\bibitem{JinPCM2016}
Jin, X., Wu, Z., Song, C., Zhang, C., Li, X.: 3d point cloud encryption through
  chaotic mapping. In: Advances in Multimedia Information Processing - {PCM}
  2016 - 17th Pacific-Rim Conference on Multimedia, Xi'an, China, September
  15-16, 2016, Proceedings, Part {I}. pp. 119--129 (2016).
  \doi{10.1007/978-3-319-48890-5_12},
  \url{http://dx.doi.org/10.1007/978-3-319-48890-5_12}

\bibitem{JinWCSP2016}
Jin, X., Yin, S., Li, X., Zhao, G., Tian, Z., Sun, N., Zhu, S.: Color image
  encryption in ycbcr space. In: 8th International Conference on Wireless
  Communications {\&} Signal Processing, {WCSP} 2016, Yangzhou, China, October
  13-15, 2016. pp.~1--5 (2016). \doi{10.1109/WCSP.2016.7752646},
  \url{https://doi.org/10.1109/WCSP.2016.7752646}

\bibitem{JinMTA2017}
Jin, X., Yin, S., Liu, N., Li, X., Zhao, G., Ge, S.: Color image encryption in
  non-rgb color spaces. Multimedia Tools Appl., First Online: 05 September 2017
   (2017)

\bibitem{JinArXiv2017}
Jin, X., Zhu, S., Xiao, C., Sun, H., Li, X., Zhao, G., Ge, S.: 3d textured
  model encryption via 3d lu chaotic mapping. Science China Information
  Sciences (SCIS)  \textbf{60}(12) (2017),
  \url{https://link.springer.com/article/10.1007/s11432-017-9266-1}

\bibitem{JolfaeiTIFS2015}
Jolfaei, A., Wu, X., Muthukkumarasamy, V.: A 3d object encryption scheme which
  maintains dimensional and spatial stability. {IEEE} Trans. Information
  Forensics and Security  \textbf{10}(2),  409--422 (2015).
  \doi{10.1109/TIFS.2014.2378146},
  \url{http://dx.doi.org/10.1109/TIFS.2014.2378146}

\bibitem{JolfaeiPsivt2015}
Jolfaei, A., Wu, X., Muthukkumarasamy, V.: A secure lightweight texture
  encryption scheme. In: Image and Video Technology - {PSIVT} 2015 Workshops -
  {RV} 2015, {GPID} 2013, {VG} 2015, {EO4AS} 2015, {MCBMIIA} 2015, and {VSWS}
  2015, Auckland, New Zealand, November 23-27, 2015. Revised Selected Papers.
  pp. 344--356 (2015). \doi{10.1007/978-3-319-30285-0_28},
  \url{http://dx.doi.org/10.1007/978-3-319-30285-0_28}

\bibitem{EluardCORESA2013}
Éluard, M., Maetz, Y., Doërr, G.: Geometry-preserving encryption for 3d
  meshes. In: COmpression et REprésentation des Signaux Audiovisuels (CORESA).
  pp. 7--12 (2013)

\bibitem{ReyHAIS2015}
del Rey, {\'{A}}.M.: A method to encrypt 3d solid objects based on
  three-dimensional cellular automata. In: Hybrid Artificial Intelligent
  Systems - 10th International Conference, {HAIS} 2015, Bilbao, Spain, June
  22-24, 2015, Proceedings. pp. 427--438 (2015).
  \doi{10.1007/978-3-319-19644-2_36},
  \url{http://dx.doi.org/10.1007/978-3-319-19644-2_36}

\bibitem{Shannon1949}
Shannon, C.: Communication theory of secrecy systems. Bell System Technical
  Journal, Vol 28, pp. 656-715  (Oktober 1949)

\bibitem{ShiACM2006}
Shi, W., Lee, H.S., Yoo, R.M., Boldyreva, A.: A digital rights enabled graphics
  processing system. In: Proceedings of the 21st {ACM} {SIGGRAPH/EUROGRAPHICS}
  symposium on Graphics hardware, Vienna, Austria, September 3-4, 2006. pp.
  17--26 (2006). \doi{10.2312/EGGH/EGGH06/017-026},
  \url{https://doi.org/10.2312/EGGH/EGGH06/017-026}

\bibitem{VermaJIPS2013}
Verma, O.P., Nizam, M., Ahmad, M.: Modified multi-chaotic systems that are
  based on pixel shuffle for image encryption. {JIPS}  \textbf{9}(2),  271--286
  (2013). \doi{10.3745/JIPS.2013.9.2.271},
  \url{https://doi.org/10.3745/JIPS.2013.9.2.271}

\bibitem{Zhen2015}
Zhen, P., Zhao, G., Min, L., Jin, X.: Chaos-based image encryption scheme
  combining {DNA} coding and entropy. Multimedia Tools Appl.  \textbf{75}(11),
  6303--6319 (2016). \doi{10.1007/s11042-015-2573-x},
  \url{https://doi.org/10.1007/s11042-015-2573-x}

\end{thebibliography}
%

\end{document}